\documentclass[conference]{IEEEtran}
\IEEEoverridecommandlockouts
\usepackage{cite}
\usepackage{amsmath,amssymb,amsfonts}
\usepackage{algorithmic}
\usepackage{graphicx}
\usepackage{textcomp}
\usepackage{xcolor}

\def\BibTeX{{\rm B\kern-.05em{\sc i\kern-.025em b}\kern-.08em
    T\kern-.1667em\lower.7ex\hbox{E}\kern-.125emX}}
\usepackage{tabularx}
\usepackage{makecell}
\usepackage{multirow}

\makeatletter
\newcommand{\linebreakand}{%
  \end{@IEEEauthorhalign}
  \hfill\mbox{}\par
  \mbox{}\hfill\begin{@IEEEauthorhalign}
}
\makeatother

\begin{document}

\title{Corporate Credit Rating: A Survey
\thanks{Identify applicable funding agency here. If none, delete this.}
}

\author{
\IEEEauthorblockN{Bojing Feng}
\IEEEauthorblockA{\textit{CRIPAC \& NLPR \& CASIA}\\
Beijing, China \\
bojing.feng@cripac.ia.ac.cn}
\and
\IEEEauthorblockN{Xi Cheng}
\IEEEauthorblockA{\textit{CRIPAC \& NLPR \& CASIA}\\
Beijing, China \\
chengxi21@mails.ucas.ac.cn}
\and
\IEEEauthorblockN{Dan Li}
\IEEEauthorblockA{\textit{Tianjin Academy for} \\
\textit{Intelligent Recognition Technologies}\\
Tianjin, China \\
lidan0320@163.com}
\\
\linebreakand
\IEEEauthorblockN{Zeyu Liu}
\IEEEauthorblockA{\textit{Tianjin Academy for} \\
\textit{Intelligent Recognition Technologies}\\
Tianjin, China \\
zeyuliu@outlook.com}
\and
\IEEEauthorblockN{Wenfang Xue}
\IEEEauthorblockA{\textit{CRIPAC \& NLPR \& CASIA}\\
Beijing, China \\
wenfang.xue@ia.ac.cn}
}
\maketitle
\begin{abstract}
Corporate credit rating (CCR) plays a very important role in the process of contemporary economic and social development. How to use credit rating methods for enterprises has always been a problem worthy of discussion. Through reading and studying the relevant literature at home and abroad, this paper makes a systematic survey of CCR. This paper combs the context of the development of CCR methods from the three levels: statistical models, machine learning models and neural network models, summarizes the common databases of CCR, and deeply compares the advantages and disadvantages of the models. Finally, this paper summarizes the problems existing in the current research and prospects the future of CCR. Compared with the existing review of CCR, this paper expounds and analyzes the progress of neural network model in this field in recent years.

\end{abstract}


\section{the Research Background of Corporate Credit Rating}
\subsection{Definition}
Up to now, there is no unified concept of corporate credit rating (CCR). It was originally derived from bond rating. The rating object is the willingness and ability of the enterprise to perform its obligations on time according to the contract. The purpose of rating is to evaluate the default risk of the enterprise as a debtor. The rated enterprises are mainly divided into non-financial companies (such as industrial companies, transportation companies and tourism companies) and financial companies (such as insurance and securities companies). Due to the particularity of capital flow and organizational structure, the credit risk of the latter is often greater and the rating work is harder.
\subsection{Relevant Rating Agencies}
At present, there are three most famous credit rating agencies in the world, namely S \& P, Moody's and Fitch. Standard \& Poor's (S \& P) has a history of more than 160 years. It plays a leading role in the industry and covers 126 countries and regions around the world. S \& P entered China's credit rating market in 2019 and updated sovereign credit ratings of various countries and regions weekly. Moody's has a history of more than 120 years which was split from Deng Baishi. Its rating covers more than 100 countries and regions around the world and entered the Chinese market in 2001. Fitch has a history of more than 100 years. Due to its short establishment time, the scale is slightly smaller than that of S \& P and Moody's. Fitch's rating range covers more than 70 countries and regions around the world and is allowed to enter the Chinese market in 2020. In addition to these three American institutions, CHinese rating agencies are also developing, such as Dagong Global Credit Rating, CCXI, Golden Credit Rating, etc. However, for the reason that Chinese rating agencies established later, there is still much work to do about rating methods. Experts and scholars use a variety of datasets when building credit rating models. Table\ref{tab1} shows some commonly used datasets.

\begin{table*}[htbp]
\renewcommand\arraystretch{1.5}
\caption{Database commonly used for enterprise credit rating}
\begin{center}
\begin{tabular}{|p{0.2\textwidth}<{\centering}|p{0.7\textwidth}|}
\hline
\textbf{Database}&\makecell[c]{\textbf{Introduction}} \\
\hline
Reset&A comprehensive data platform in China, which provides financial markets data all over the world. It mainly includes stocks, gold, research reports, macro statistics and other series\\
\hline
Financial report of listed company & A report on the disclosure of information such as business profile and financial status of listed company\\
\hline
The credit investigation system of the people's Bank of China & The most complete enterprise and individual credit investigation database in China\\
\hline
CSMAR & A research-based database in the field of economy and finance according to China's national conditions, established in 2001. It covers 18 series, such as factor research, green economy, stocks, information and funds. It includes 160 + databases, more than 4000 tables and more than 50000 fields.\\
\hline
WRDS & A cross database research tool in the financial field developed by Wharton School of business, University of Pennsylvania in 1993, integrates many famous databases such as Compustat, CRSP, TFN and TAQ\\
\hline
Bloomberg & The world's largest provider for financial information and financial data services\\
\hline
FAME & covers 3.8 million companies in the UK and Ireland\\
\hline
UCI machine learning knowledge base & A database for machine learning proposed by the University of California Irvine\\
\hline
Compustat & Released by standard \& Poor's, includes the financial data of Listed Companies in North America and around the world for nearly 20 years\\
\hline
CRSP & An authoritative database of Listed Companies in the field of securities and exchange, established by the University of Chicago\\
\hline
NEEDS & Japan's largest comprehensive economic database\\
\hline
Bankscope & A banking information database developed by BVD in cooperation with FitchRatings. As an authoritative rating agency in the banking industry, it provides the operation and credit analysis data of more than 12800 major banks and important financial institutions in the world.\\
\hline
WIND & A financial and securities database provided by a Chinese enterprise Wind\\
\hline
Kis-value & A database in South Korea that provides business reports and stock market data analysis\\
\hline
\end{tabular}

\label{tab1}
\end{center}
\end{table*}

\subsection{The significance of enterprise credit rating CCR}
As a social intermediary service under the condition of market economy, enterprise credit rating plays an important role in maintaining social and economic order. When developing customers, enterprises should be based on a comprehensive understanding of customers' credit status, blindly pursuing a large number of orders will inevitably receive bad debts. Credit rating can help financial practitioners avoid risks, provide investors and partners with objective and fair credit information, and reduce the management pressure of enterprises. In order to maintain the stability of capital order, the capital market management department needs to conduct regular investigate of enterprises. Enterprise credit rating improves the supervision ability of economic management departments and promotes the social popularity of enterprises.

Similar to corporate credit rating is bond credit rating. Although both of them are to promote the effective allocation of resources and reduce the information asymmetry between rating objects and investors, their rating objects are different. The former is for commercial banks or relevant regulators, and the latter is for bond investors. Due to the uniqueness of bond issuance, the credit rating of enterprises is high, and the bond rating issued by enterprises may not be high.  In addition, for the reason that bond rating agencies take a long time and go through multiple stages such as information collection, processing, secondary rating and tracking rating, the change of credit risk is often not responded in time. It is very important for financial institutions to establish their own credit rating system. However, a complete and detailed rating by rating agencies requires a lot of human resources, capital and time \cite{Huang2004CreditRA}, which is difficult for most enterprises and financial practitioners to bear. Therefore, it is certainly valuable for investors to build an accurate enterprise credit rating model.

At present, many experts and scholars have done detailed research on enterprise credit rating models, but the overall survey of such models is relatively insufficient. There are few reviews of rating models based on neural networks in recent years. This paper combines the traditional statistical models, the machine learning models and the neural network models with a high topic in recent years, to make a comprehensive survey on corporate credit rating.





\begin{table}[htbp]
\caption{Common indicators of corporate credit rating}
\begin{center}
\begin{tabular}{|p{0.1\textwidth}<{\centering}|p{0.3\textwidth}<{\centering}|}
\hline
\textbf{Category}&\textbf{Indicator} \\
\hline
\multirow{8}{*}{\textbf{Size}}& Total assets\\
\cline{2-2} 
& Total capital\\
\cline{2-2} 
& Total cash flow\\
\cline{2-2} 
& Total equity\\
\cline{2-2} 
& Total capital expenditures\\
\cline{2-2} 
& Market capitalization\\
\cline{2-2} 
& Total trading volumes\\
\cline{2-2} 
& Total debt\\
\hline
\multirow{7}{*}{\textbf{Profitability}}& Earnings before interest and tax\\
\cline{2-2} 
& Earnings after tax\\
\cline{2-2} 
& Net income\\
\cline{2-2} 
& Sales to net worth\\
\cline{2-2} 
& Sales to total assets\\
\cline{2-2} 
& Non-cash working capital\\
\cline{2-2} 
& Sales of previous year\\
\hline
\multirow{6}{*}{\textbf{Equity market}}& Dividends\\
\cline{2-2} 
& Growth in EPS\\
\cline{2-2} 
& Price to book value ratio\\
\cline{2-2} 
& Retained earnings\\
\cline{2-2} 
& Stock price\\
\cline{2-2} 
& Number of shares outstanding\\
\hline
\end{tabular}
\label{Common indicators of corporate credit rating}
\end{center}
\end{table}

\section{Corporate credit rating based on different models}
\subsection{Statistical models}
Most of the traditional models used to evaluate corporate credit risk in the financial field are based on statistical scenarios and derived from the corporate bankruptcy prediction model. These models build a rating index system based on enterprise financial indicators(such as asset liability structure, cash flow, profitability, asset liquidity, etc.), and then use statistical methods to analyze the features of these index systems to complete the classification of enterprise credit rating. The commenly used indicators are listed in table \ref{Common indicators of corporate credit rating}.The following are the commonly used statistical models.

\subsubsection{ZETA}
Zeta model was originally used to evaluate the risk of corporate bankruptcy and had also been widely used in the field of credit rating,later. The evaluation index system is constructed based on seven dimensions: return on assets, income stability, debt solvency, cumulative profitability, liquidity, capitalization degree and scale. The work\cite{74-1977ZETA} combines the discriminant analysis algorithm and introduces prior probability to analyze the credit value. However, the seven indicators of zeta model are fixed and cannot cover all rating elements. After that, the widely used feature engineering is essentially looking for factors indicators suitable for describing rating problems.
\subsubsection{AHP} Analytic Hierarchy Process (AHP) refers to the method of decomposing the decision-making process into multiple levels for qualitative and quantitative analysis, which is widely performed in credit rating\cite{79-2005Development}. At present, it is used to determine the importance order of parameters. In order to balance the subjectivity of AHP method, the work\cite{75-2017A} combined objective DEA to evaluate the credit level. Analytic hierarchy process has obvious advantages for complex systems that are difficult to be fully quantitatively analyzed. However, analytic hierarchy process requires decision makers to compare the indicators and judge their relative importance by themselves. It not only introduces more subjective elements, but also is difficult to judge which one is more important when there are a large number of indicators.
\subsubsection{MDA}
Multiple discriminant analysis (MDA) is a statistical method to analyze the category of new data according to the original data set. It has been widely used in credit scoring model. Discriminant methods can be divided into distance discriminant method, Fisher discriminant method, Bayes discriminant method, step-by-step discriminant method, etc. Reichert et al.\cite{81-0An} pointed out that most MDA models assume that variables are distributed in multivariate normal distribution. If the true distribution of the data deviates significantly from the normal, the classification results will be seriously wrong. The Z-score model was created based on MDA. It was initially used for enterprise bankruptcy analysis and later used for credit risk measurement\cite{82-1997Credit}. Compared with the first two methods, we can find that MDA method considers the scientificity of data more and is less affected by the subjective attitude of analysts. However, MDA classification results are more accurate when the data set is small. In addition ,the data which are statistically independent of the estimated samples are not used in MDA model to verify the accuracy of the rating, resulting in the deviation of the verification results. MDA model assumes that the variance-covariance matrix of all categories is equal, which is inconsistent with reality.
\subsubsection{MARS}
Multivariate adaptive regression splines(MARS)\cite{78-1991Multivariate} is also used for credit rating. Compared with the MDA model suitable for small data sets, MARS can accurately and quickly deal with the corporate credit rating problem with a large number. More significantly, MARS can also capture the nonlinearity and interaction between variables. However, compared with other models, MARS is not used much in enterprise credit rating.

\subsubsection{LR}
Logistic regression is a nonlinear model supported by linear regression theory, which is often used to deal with binary classification problems. Laitinen et al.\cite{83-2004Predicting} used logistic regression and linear regression models to analyze corporate credit risk. West et al.\cite{84-1985A} constructed a commercial bank early warning system by combining classical factor analysis and multivariate logit estimation. As a widely used classification algorithm, it is fast and can be used for nonlinear classification. It is often combined with other algorithms for credit rating problems.

\subsubsection{Others}
This work\cite{76-2007The} also combines qualitative and quantitative indicators to establish a  credit evaluation index system for small enterprises. The index weight is generated by the incremental clustering algorithm, which is based on the chemical recognition system of ants. Experiments show that the clustering algorithm is suitable for high-dimensional features. Shi et al.\cite{77-2018A} used Pearson correlation analysis and F test significance discrimination to screen key features of small enterprise financing ability, so as to give consideration to classification accuracy and calculation efficiency. Due to the simplicity and sorting ability of fuzzy TOPSIS method, it is also used for enterprise credit rating\cite{80-2010Development}.

Throughout the history of statistical based models, we can find that except for a few traditional models, all models are good at dealing with linear relationships rather than nonlinear relationships. Large rating agencies often pay attention to the importance of analysts' subjective judgment in determining rating model parameters or direct credit rating. Although these traditional methods are based on statistics and analysis of data, they have been affected by human factors on enterprise credit rating. In addition, traditional models (such as logit regression and KMV) often make assumptions about the parameter distribution of the data set when doing rating tasks. These assumptions may contradict the real data distribution. The machine learning model provides answers to these problems.

\subsection{Machine learning models}


\subsubsection{SVM}
SVM is one of the most commonly used models in the field of credit rating. Huang et al.\cite{Huang2004CreditRA}used SVM to conduct comparative research on credit rating analysis market, and concluded that SVM has higher accuracy than logistic regression model. They found that using a small and accurate indicator data set to rate credit results were even more accurate than financial data sets containing a wide range of indicators. The work also compares and analyzes the data indicators concerned by rating agencies in the United States and Taiwan. It is found that the former pays more attention to the size of the company and the latter pays more attention to the profitability of the company. This coincides with the way American companies advocate high leverage operation. Different from the traditional statistical learning model, SVM emphasizes the objectivity of rating methods, and financial variables can determine the rating results better than institutional analysts. The application of feature engineering can often improve the accuracy of SVM, which raises the upper limit of SVM, but also limits the development of SVM in the field of credit rating.

SVM was originally designed for binary classification. Simple binary classification is not suitable for enterprise credit rating because credit rating is not absolutely good or bad for enterprises. Fuzzy SVM\cite{66-Yongqiao2005A} assigns different memberships to each sample of positive and negative categories, which enhances the generalization ability of SVM. Then, with the development of multi-class support vector machines (MSVMs),  the algorithm of "one to one", "one to all" and directed acyclic graph SVM (DAGSVM) are used in corporate credit rating. This work\cite{67-2006Bond} used the above three multi-classification SVM methods for corporate bond rating, and comes to the conclusion that DAGSVM has the best performance.In order to realize the classification of nonlinear samples, RBF kernel function is used to increase the dimension of input space. This work\cite{68-2007Application} uses SVM with RBF kernel function to classify enterprise credit rating. The optimal parameter value of RBF kernel function is found by grid search technology.  Experiments\cite{66-Yongqiao2005A} showed that SVM with appropriate kernel and membership generation method has higher accuracy in credit rating than standard SVM and fuzzy SVM.

However, enterprise credit rating is not a simple classification problem, and there is a sequential relationship between the classification results. Considering the unique order of credit rating, SVM based on ordered pairwise partition (OPP) strategy is proposed\cite{69-2012A}. In order to solve the problem of model parameter optimization, genetic algorithm is also introduced into SVM. GA-SVM\cite{A2007Credit} uses grid search to set model parameters and genetic algorithm to optimize parameters. But SVM is always a black box problem. Understanding the principle of SVM will improve the practicability of the model. In order to improve the interpretability of SVM, CRCR-SVM\cite{70-2013Rule}is trained through the covering reduction algorithm inspired by the traditional rule learning methods.

Besides, SVM is often used for feature selection. The process of feature selection is very tortuous,Due to the interdependence between indicators. Corporate credit rating depends on multivariable factors. Fisher is a commonly used feature selection method. This method is easy to implement and fast, but does not take into account the interaction between variables and classifiers. Other two-sample independence metrics, such as KS and chi-squared tests, usually get results similar to Fisher's score. In order to avoid the additional calibration steps required by the feature ranking method, Maldonado et al.\cite{65-2017Cost-based} used two SVM methods (L1-MISVM, LP-MISVM) to obtain the best feature subset of bank credit loan data. When constructing the classifier, this method will consider the interaction of all variables and the relationship between them. The work\cite{67-2006Bond} proposed that feature selection can further improve the generalization performance of SVM. Compared with genetic programming and decision tree classifier, the input accuracy of SVM is less dependent on a large number of input features\cite{A2007Credit}.

\subsubsection{Decision tree}
Decision tree (DT) is an interpretable algorithm that can be constructed quickly. Both ID3 and gdbt algorithms are used in corporate credit rating, and the latter is more effective\cite{Gu2019Duration}. Combining the method of ensemble learning and decision rules, the correlation-adjusted decision forest (CADF)\cite{73-2015Enhancing} balances the accuracy and interpretability of the model. This method uses the decision tree as the base classifier and selects 18 most important features for credit risk rating. Moody's KMV model (KMV) is a famous credit rating analysis model based on financial theory and default probability. The hybrid KMV model\cite{72-2012A} combines KMV with random forests (RF) and rough set theory (RST) to improve the accuracy of credit rating. RST does not make any assumptions about the distribution of data and is suitable for quantitative and qualitative analysis. When the decision-making process involves uncertain fuzzy data, RST has achieved remarkable results in solving decision support problems. First, the work use KMV to predict variables, then RF selects the variables as the input of RST model. Finally, the model generates results in the form of if-then rules. The process is transparent and easy for decision makers to understand. However, this method must optimize some parameters to construct the base classifier of RF. However, due to the high dimension, sparsity and strong correlation of financial data, the performance of DT model is limited. In subsequent studies, DT is often used as the baseline of credit rating model or combined with other algorithms. 

\subsubsection{Ensemble learning}


Ensemble learning is not a single machine learning algorithm, but the idea of combining multiple base learners to complete the learning task together. Due to the outstanding performance of ensemble learning in classification tasks, the algorithm has also made great progress in CCR. The basic idea of ensemble learning is that when we make important decisions, we need to refer to various opinions from different aspects.  Decisions made by weighting opinions from multiple perspectives are often more appropriate than decisions made by referring only to unilateral opinions. Therefore, in addition to having a certain accuracy, basic learners should have diversity with each other. When the opinions given by the base learner are complementary, the integrated learning algorithm can usually achieve better results. The Credal Decision Tree (CDT) proposed by Abellan and Castellano\cite{46-2017A} changes the way of dealing with imprecision. They used imprecise probability and uncertainty measures to build the model, and made the base classifiers relatively unstable. Instability means that a small change in training data will make a large difference in the model, resulting in the diversity of base classifiers, which is very suitable for ensemble learning.


In recent years, neural network method combined with ensemble learning has been applied in CCR task by many experts. Donate proposed that in the CCR task, the model integrating multiple neural networks is better than using only a single neural network. The ensemble strategy of the fitness weighted multiple base learners is more accurate than the non-weighted strategy, which proves the great potential of the neural network method using the ensemble algorithm\cite{45-2012Time}. In order to ensure the difference between base learners, the multistage neural network ensemble learning model proposed by Yu et al.\cite{44-2008Credit} adopted the the decorrelation maximization method when selecting ANN, and then scaled the decision value of each single base learner from $(-\infty,+\infty)$ to $(0,1)$. Then the base classification results were integrated using maximum, minimum, median, average and product strategies respectively. Among them, the product strategy has the best performance, followed by the mean strategy. The reason for this result remains to be discussed.

\begin{figure}[htbp]
\centerline{\includegraphics{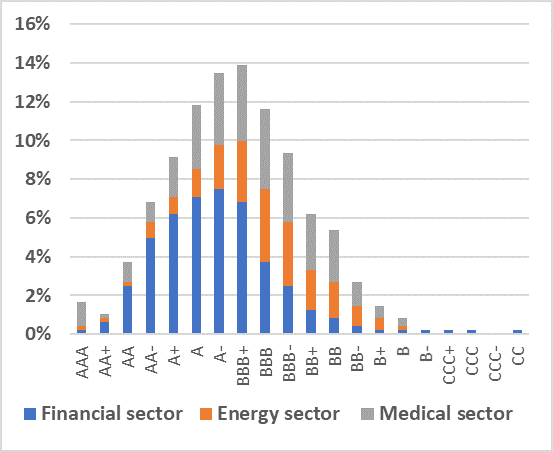}}
\caption{Distribution chart of CCR}
\label{Distribution chart of CCR}
\end{figure}
\begin{figure}[htbp]
\centerline{\includegraphics[width=8cm]{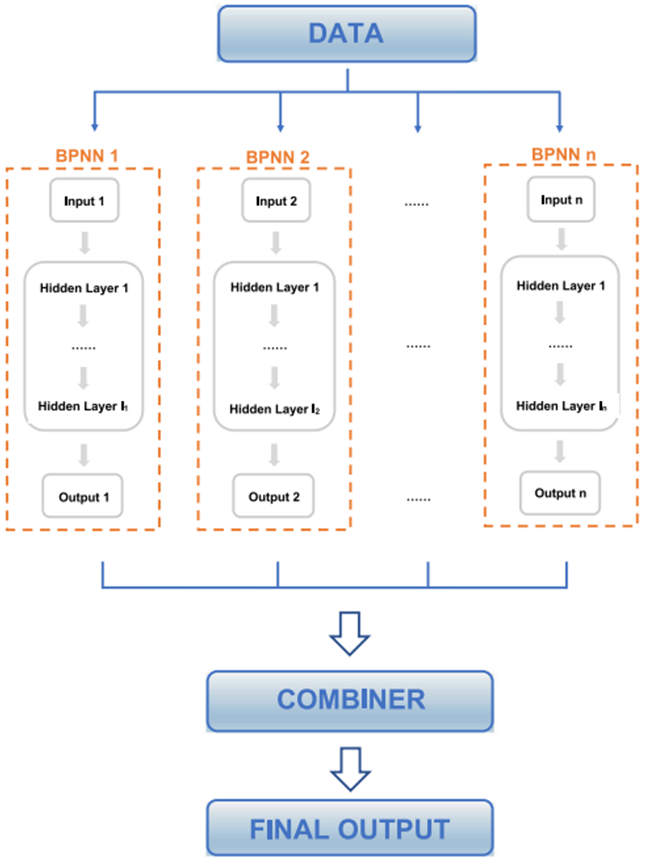}}
\caption{Ensemble Learning based PANNs framework}
\label{Ensemble Learning based PANNs framework}
\end{figure}

The commonly used algorithms for ensemble learning include majority voting, weighted average, Bagging, Boosting, Random Subspace, DECORATE, Rotation Forest, etc. The majority voting algorithm is most commonly used, but it ignores the fact that minorities sometimes produce correct results. The majority voting algorithm regards the confidence of each neural network as the same. In the CCR task, the available sample data is often limited, while bagging algorithm generates more estimates for the overall dataset in a statistical way by randomly sampling the subset from the original training set. Sampling makes up for the limitation of training data to a certain extent, and can usually improve the classification effect\cite{44-2008Credit}. In addition, the categories of enterprise credit rating data are seriously unbalanced. As shown in the figure \ref{Distribution chart of CCR}, there are relatively few high or low ratings. Brown and Mues proposed random forest (RF) and gradient boosting algorithm, which have good performance in alleviating this problem. However, He et al. proposed that when solving the problem of unbalanced enterprise credit rating data, the parameters of RF or gradient boosting model need to be further optimized\cite{47-2018a}. They extended the BalanceCascade method, generated adjustable balance subsets based on the imbalance rate of training data, and constructed a three-stage integration model with RF and XGBoost as base classifiers. The three stage integration model uses the prediction results of the previous layer as the new interpretation features of the next layer, and the parameters of the base classifier are optimized by particle swarm optimization. However, the choice of EBCA threshold remains to be explored. Feature engineering is also used in artificial neural networks ensemble model. Considering the impact of historical financial data and credit rating on the current credit rating, Wang et al. \cite {18-wang2021Utilizing} constructed the parallel artificial neural networks (PANNs) based on ensemble learning, and its classic and concise framework is shown in figure \ref{Ensemble Learning based PANNs framework}. They expanded the feature space, in which the disparity and first-order variation can predict the CCR score better.

However, how to decide the number and form of base classifiers in ensemble learning method and make it more suitable for enterprise credit rating remains to be studied. Ensemble learning integrates many base classifiers, and its training time also increases exponentially. The balance between the number of base classifiers and training time is worth weighing in the future. Although the model combined with multiple base classifiers can consider problems from many aspects, the integration strategy is also difficult to be suitable for all multiple base classifiers at the same time. How to improve the performance of the model containing multiple base classifiers is the research direction of ensemble learning methods in the future. The experimental results of Wang et al.\cite {18-wang2021Utilizing} show that sometimes the accuracy of neural network-based integration model does not increase significantly with the increase of data volume. One possible explanation is that more financial data will introduce more noise. In the future research, we can get more useful datasets to improve the prediction accuracy by reducing noise.

\subsubsection{Others}
With the purpose of improving the interpretability of the model, the work\cite{71-2013Hybrid} is based on knowledge systems to generate decision rules, and uses rough sets (RSs) to construct two hybrid models to classify the credit rating of banking industry. Chai et al.\cite{6-chai2019A} conducted credit rating for small enterprises and found that the impact of non-financial factors is greater than financial factors for them. When constructing the credit index system, they first use triangular fuzzy numbers to convert the qualitative data into numerical values, and then use partial correlation analysis (PCA) and probit regression algorithm to eliminate redundant indexes. In the classification stage, this work uses TOPSIS algorithm to calculate the credit score, and then uses fuzzy C-means (FCM) to cluster the credit score of small enterprises.

\subsection{Neural network models}







Looking at the machine learning models, we can find that although these models are more objective than the previous statistical models, they only use quantitative data and discard the information brought by qualitative data to a certain extent. In the problem of enterprise credit rating, qualitative data such as text contain a lot of risk information. In addition, such models rely too much on Feature Engineering, and good features will significantly improve the effect of the model. The neural network model provides a good answer to these questions.

\subsubsection{Early NN}
Before neural network was widely used, feature engineering has always been the focus of financial engineering, especially in corporate credit rating. In the previous machine learning algorithms, the rating results are more accurate when using the new features obtained by feature engineering than using the original features. Chen et al.\cite{3-2020A} used the statistical method of one-way ANOVA to select features, which revealed that although the statistical method will improve the classification accuracy on the training set, it also brings noise and leads to over fitting. However, for the reason that NN gives different weights to features during training, important features selected as the focus items. Golbayani et al. \cite{12-golbayani2020Application}found that the rating results are more accurate when using all financial variables as inputs and performing feature engineering in the training process of NN. Compared with statistical methods and machine learning methods, NN does not assume the data distribution. 

Early NN (such as multilayer perceptron) need to adjust the learning rate manually and avoid falling into local minimum. Compared with traditional machine learning methods, multilayer perceptron (MLP) can effectively deal with high-dimensional data and nonlinear relationships. Brennan et al.\cite{62-2004Corporate} used the information in the financial statements to build a back propagation neural network (BPNN) to rate the bond issuing companies. They found that the results obtained using BPNN were much more accurate than traditional statistical methods. Huang et al.\cite{Huang2004CreditRA} used a BPNN model to explain CCR, and tried to analyze the importance of different input financial variables from the model. Angelini et al.\cite{64-2008A} believe that data analysis and processing and parameter optimization are the key and difficult points to solve the company's credit rating. They use a classical feedforward neural network and a special feedforward neural network with ad hoc connections to evaluate credit risk. The latter consists of four layers of feedforward network. Each group consists of three neurons connected to the next input layer.

Nevertheless, the convergence speed of MLP model is slow and the process is unstable. MLP is more suitable for binary classification but has lower accuracy in multi class classification. In addition, the traditional BPNN algorithm trains a large number of parameters in network, resulting in over fitting easily and a long training time.The work\cite{Huang2004CreditRA,A2007Credit,68-2007Application,69-2012A} suggested that NN is not as good as SVM in credit rating. Choi et al. argued that it may be because SVM is more robust in avoiding over fitting problems, and there are fewer parameters compared with NN\cite{1-2020Predicting}. Du et al.\cite{7-2018Enterprise}argued that genetic algorithm can modify and optimize the parameters of neural network and improve the accuracy of enterprise credit rating. The genetic algorithm credit rating model alleviates the problems of long training time, slow convergence speed and the possibility of falling into local minimum of BP neural network to a certain extent.

Many other neural network models have been proposed one after another based on the classical feedforward network model on the issue of CCR. The deep network neural structure (DNN) is composed of multiple shallow neural networks. With the increase of the number of network layers, the gradient of DNN will vanish in training, the optimization function is more and more easy to fall into the local optimal solution, and the training efficiency is greatly reduced. Until 2006, Hinton proposed the method of training Restricted Boltzmann Machine layer by layer, which improved the above problems. This structure is called Deep Belief Network (DBN). Luo et al. did the first research on enterprise credit scoring using DBN\cite{luo2017deep}. The classification performance is better than Multinomial Logistic Regression (MLR), Multilayer Perceptron (MLP) and SVM. In addition, Kim et al.\cite{61-2005Predicting} used adaptive learning networks (ALN) to predict bond ratings. The work\cite{63-hajek2010Probabilistic} uses PNNs to rate U.S. companies and municipalities. Correlation based approach and genetic algorithms are used in the data preprocessing stage. They found that probabilistic neural networks have more accurate results than other benchmark classifiers, such as NNs (FFNN, RBF, data processing polynomial neural network), cascaded correlation neural network and statistical methods (LR, MDA). Probabilistic neural network is a branch of radial basis function network, which belongs to a kind of feedforward network. It combines density function estimation and Bayesian decision theory to make the judgment interface close to the Bayesian best judgment interface. It has the advantages of simple learning process and fast training speed.

\subsubsection{CNN}
CNN came into being in order to solve the problem of parameter explosion of DNN, which has been proved to be significantly superior to the traditional machine learning technology in various financial problems, especially in the field of stock market analysis\cite{57-2016Credit,58-2019Convolutional,59-2017Classification-based}.CNN is mainly composed of the convolution layers, the pooling layers and the fully connected layers and training by back propagation algorithm. As far as we know, Golbayani et al. \cite{12-golbayani2020Application} is the first to use CNN for corporate credit rating. They proposed CNN model and CNN2D model with dropout and early stopping algorithms to tackle the corporate credit rating problem. The two models are all composed of two convolution layers and two fully connected layers. The difference between them is that the filter moves only in one direction in CNN model but two in CNN2D model. The work also proposed a two-way ANOVA model which compares the multiple performance of the network architectures. Besides, not all companies could provide credit rating scores or financial information in each year. Compared with MLP, CNN achieves good results in dealing with the problem of missing data due to its efficiency in adjusting the feature weight.

\begin{figure*}[htbp]
\centerline{\includegraphics[width=19cm]{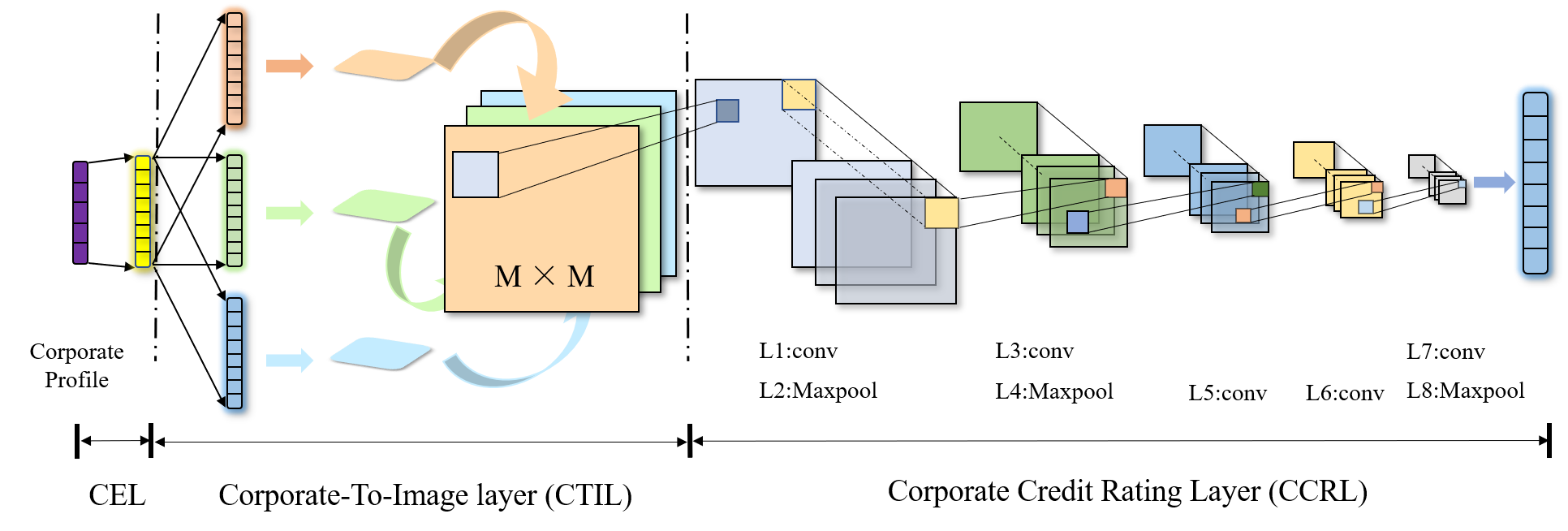}}
\caption{The architecture of CCR-CNN\cite{56-2020Every}}
\label{CCR-CNN}
\end{figure*}

Various derived CNN models are widely used in CV and NLP fields, but they are not suitable for financial scenarios. The early NN models only extracted one-dimensional features of enterprise data. Inspired by the CV field, Feng et al.\cite{56-2020Every} constructed a CCR-CNN model to generated an image containing two-dimensional financial information for each enterprise, which was fed into the CNN structure and obtained the classification results. The CCR-CNN model architecture is shown in figure \ref{CCR-CNN}. The advantage of this model is that it captures the unique two-dimensional relationship between enterprise indicators and build graphics, which has been ignored in previous models. However, corporate credit rating is dynamic and strongly related to time factors. RNN is a good algorithm to establish a continuous enterprise credit rating model.


\subsubsection{RNN}
Although RNN has achieved remarkable achievements in the study of time series problem, the gradient disappearance and gradient explosion caused by back-propagation algorithm has brought great difficulties to the training of RNN. Long short term memory network is a variant of RNN, which combines short-term memory with long-term memory through a gating mechanism, and solves the problem of gradient explosion and disappearance to some degree. However, LSTM has high computational complexity and exquisite structure. The gated recurrent unit (GRU) 
of RNN can save computing cost while ensuring considerable accuracy. Attention mechanism is proposed to further save computing cost. The principle of attention mechanism is to select the information more critical to the current task from a large number of information by assigning weights. The attention distribution is realized by calculating the similarity or correlation of vectors. At present, most attention models are attached to the encoder decoder framework. For example, the parallel computing of the encoder decoder construction named transformer 
can significantly reduce the training time. In addition, some models such as Bert 
, deep transformer 
, transformer XL 
have also been proposed.

\begin{figure}[htbp]
\centerline{\includegraphics[width=9.5cm]{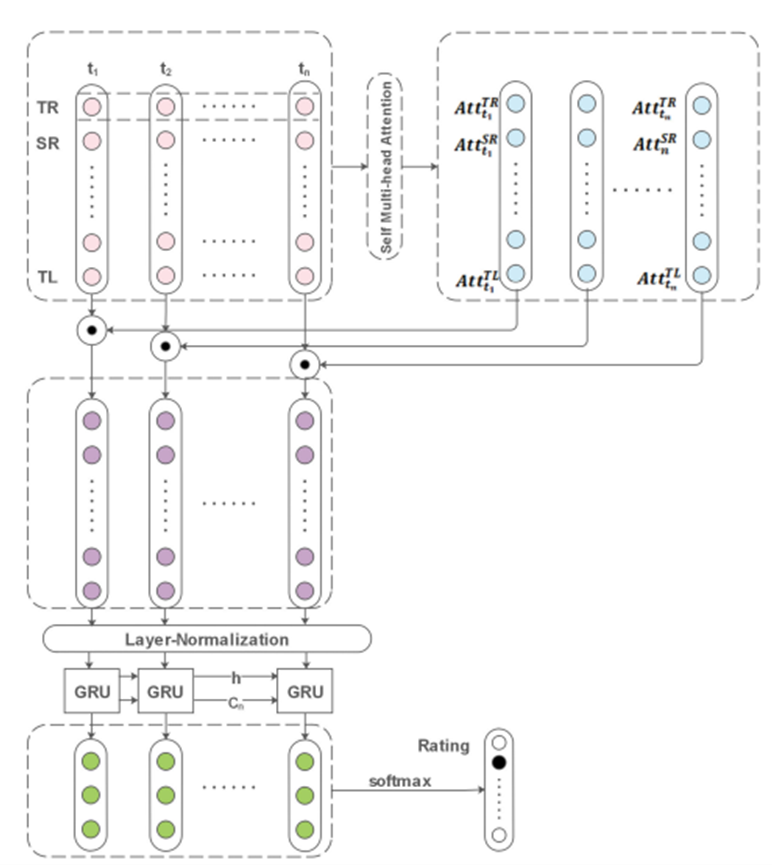}}
\caption{The architecture of SMAGRU \cite{3-2020A}}
\label{SMAGRU}
\end{figure}

Golbayani et al.\cite{12-golbayani2020Application}compared CNN with LSTM and proved that the latter is better in dealing with corporate credit rating issues by experiment analysis. They construct the model with 32 LSTM cells and two fully connected layers. SMAGRU\cite{3-2020A} is the first work to realize the long attention mechanism in corporate credit rating. The architecture is based on GRU with multi-head self-attention mechanism, which can capture the characteristics of time series. SMAGRU is stacked by six same modules, each is composed of multi-head self-attention mechanism and fully connected feed-forward network. The architecture of SMAGRU is shown in figure \ref{SMAGRU}. Multi-head attention is similar to multiple filters in CNN, which helps to capture more comprehensive information. Multi-head self-attention mechanism improves the classification accuracy and convergence speed by the time characteristics enhancement. 
In addition, such mechanism along with gated recurrent neural network can also adapt well to high-dimensional and sparse data, which is quite suitable for CCR.

\subsubsection{Model based on qualitative information}
Most rating methods use quantitative data(such as financial information and capital liquidity), but qualitative data (such as the company's strategic layout, public opinion and management efficiency) also have an important impact on credit rating. Credit rating is to guide investors in the future, but the financial data used is based on the company's historical operation. Moreover, financial data can not completely reflect the economic environment of enterprises. In addition, S \& P uses reports and management interviews to supplement the rating model. 

The commonly used methods to convert text into embedded vectors are BOW, Word2Vec and Doc2Vec. BOW, which was first proposed, regarded documents as a collection of words, ignoring the elements of word order, grammar and syntax. Bow assumes that each word in the document appears independently. Each dimension of the vector corresponds to the words in the corpus one by one, which indicates the importance of words. TF-IDF is the most commonly used method to calculate the relative importance of words. Word2Vec assumes that words that often appear together in the context have similar meanings, and embeds a word into a continuous vector space. The goal of training the neural network is to accurately predict the target word according to the input word. Word2Vec has two most common architectures: CBOW model and Skip-Gram model. CBOW model uses one-hot coding of adjacent words as input and predicts the word. On the contrary, Skip-Gram model uses the one-hot encoding of the word and predicts its adjacent words\cite{2013Efficient}. Doc2Vec\cite{2014Distributed} adds a paragraph vector to Word2Vec, which can be used to embed variable length text, such as sentences, paragraphs and  documents.

 Choi, J. et al. used the above three text embedding methods to obtain vectors, and fed the vectors into ANN, SVM and RF models respectively\cite{1-2020Predicting}.The experimental results show that the model trained with quantitative financial data and qualitative text data has higher accuracy than that trained with the former only. The model accuracy from low to high is Word2Vec, Doc2Vec and BOW. The reason why BOW got the highest accuracy may be that the training dataset is small. The training data has a long length, and Doc2Vec is better at handling longer documents than Word2Vec. In the future, training with larger data sets is a way to solve this problem. Feng et al.\cite{52-2020Every}performed one-hot encoding on qualitative data, and then used the embedding layer to connect quantitative financial data with it as input.

\subsubsection{GNN}
When the market of an industry is bad, the rating results of relevant enterprises tend to get worse. The relationship between enterprises is also one of the unexplored influencing factors of corporate credit rating. However, most of the existing models using graph neural network are based on the global perspective to do research on enterprises. They directly establish networks among enterprises without considering the interaction of internal characteristics of a single enterprise (such as the relationship between debt and capital structure).

\begin{figure*}[htbp]
\centerline{\includegraphics[width=18cm]{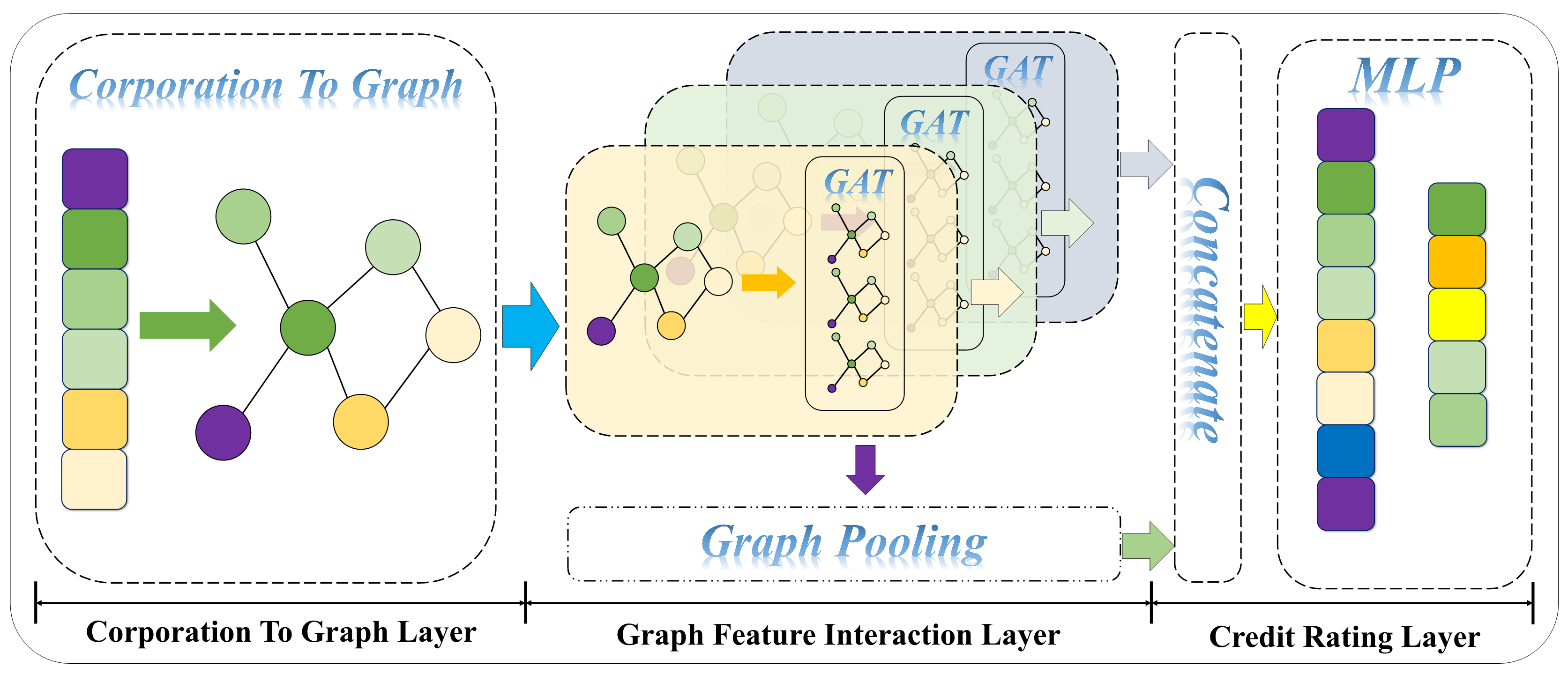}}
\caption{The architecture of CCR-GNN\cite{52-2020Every}}
\label{CCR-GNN}
\end{figure*}

CCR-GNN\cite{52-2020Every} is the first application of graph neural network to study corporate credit rating. Rather than simply regarded as a node, CCR-GNN constructs a graph for each enterprise taking into account the interaction of internal characteristics of a single enterprise. CCR-GNN includes three layers of sub neural networks. Firstly, each enterprise is mapped into a graph structure according to the relationship between features. Then these features capture local and global enterprise credit information through the interaction of graph attention layers(GAT). Finally, the credit rating layer outputs categories according to these enterprise credit information.  The model structure is shown in figure\ref{CCR-GNN}. By superimposing multiple graphics attention layers, CCR-GNN can explore high-order feature interaction in a clear way. The enterprise credit information of the characteristic node is transmitted to the adjacent nodes according to the attention mechanism.

\subsubsection{Model based on confrontation and semi-supervised learning}
The financial information of small and medium-sized enterprises is insufficient, and they do not have enough capital to support the rating. Therefore, the previous research on enterprise credit rating only considered large enterprises. However, the financial data of small and medium-sized enterprises also have research value. Semi-supervised learning algorithm provides inspiration for solving this problem. Based on the hypothesis that similar samples have similar outputs, semi-supervised learning uses both labeled data and unlabeled data to train the model. 
Adversarial learning refers to the machine learning algorithm that can resist attack based on the understanding of the attacker's ability and attack consequences. The method to realize adversarial learning is to make the two networks compete against each other, in which the generator network adds noise to the sample to construct pseudo data, and the discriminator network judges the authenticity of the data. The ability of generator and discriminator will be continuously enhanced through repeated confrontation.

\begin{figure*}[htbp]
\centerline{\includegraphics[width=19cm]{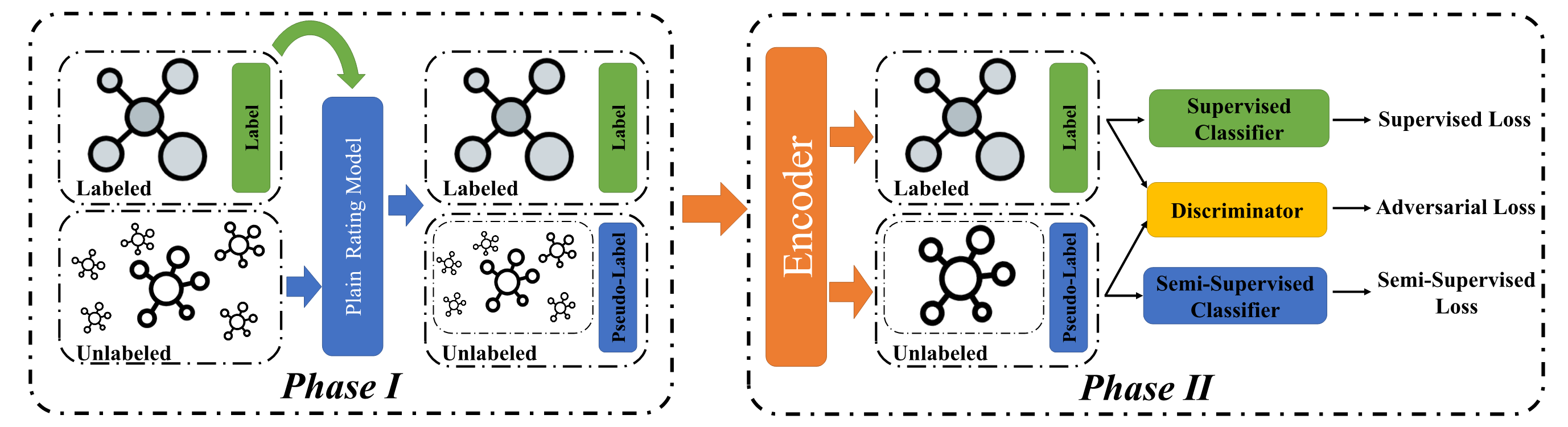}}
\caption{The architecture of ASSL4CCR\cite{42-2021-Adversarial}}
\label{ASSL4CCR}
\end{figure*}

However, Bojing Feng and Wenfang Xue find that only use semi-supervised learning will lead to the problem of representations misalignment between supervised task and semi-supervised task. Encoder module and adversarial learning are introduced in ASSL4CCR\cite{42-2021-Adversarial} to alleviate this phenomenon. ASSL4CCR include two phases which the first phase is to get the pseudo label by semi-supervised learning. In the second phase, after mapping an endocer module, the labeled data is combined with the pseudo-labeled data. The discriminator module is used to distinguish whether the data comes from true label or pseudo label. The model structure is shown in figure \ref{ASSL4CCR}.

\subsubsection{Model based on pre-training and self-supervised learning}
\begin{figure}[htbp]
\centerline{\includegraphics[width=9.5cm]{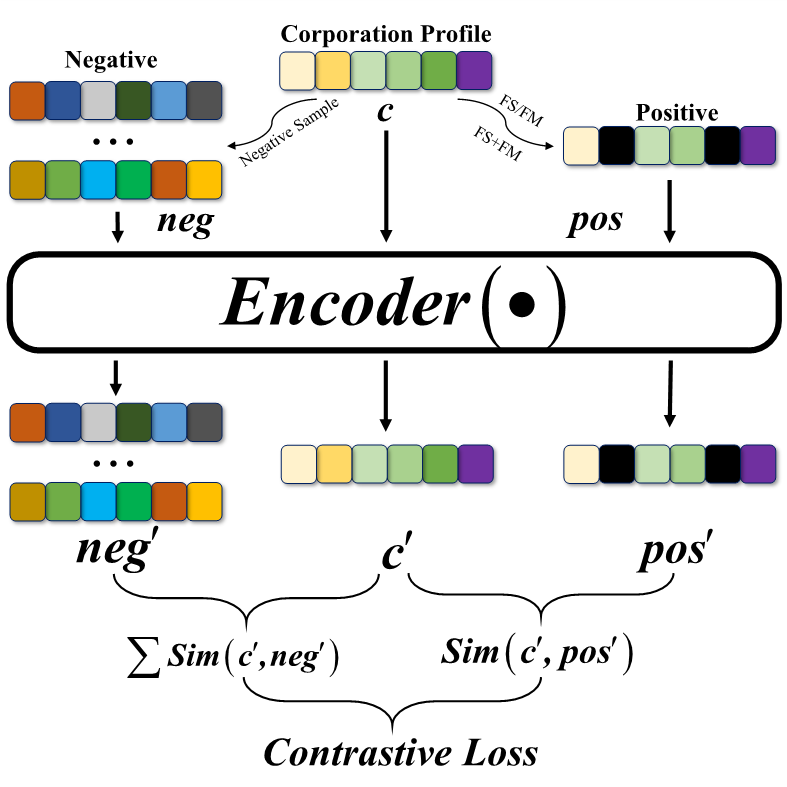}}
\caption{The architecture of the Contrastive Self-superivised Pretraining module in CP4CCR\cite{43-2021Contrastive}}
\label{CP4CCR}
\end{figure}

In addition to ensemble learning, self-supervised learning can also solve the imbalanced class problem of corporate credit dataset to a certain extent. Self-supervised learning mainly uses the pretext task to mine supervision information from unsupervised data. The valuable representation of downstream tasks is learned through constructed supervision information. CP4CCR \cite{43-2021Contrastive}adopts feature masking and feature swapping as two self-supervised tasks. Space concatenation is better than space fusion in space stacking module. After pre-training the network, transformer is a better encoder module for the standard corporate credit rating model. The framework of the contrastive self-supervised pretraining module is shown in figure \ref{CP4CCR}.

\subsubsection{Model based on interpretable learning}
\begin{figure}[htbp]
\centerline{\includegraphics[width=9.5cm]{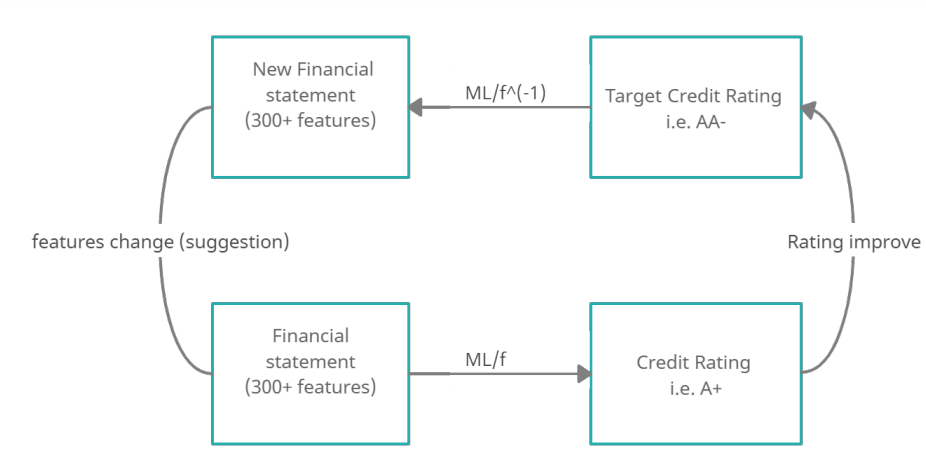}}
\caption{Example of counterfactual explanation\cite{15-wang2021A}}
\label{Example of counterfactual explanation}
\end{figure}

In the financial field, it is critical to improve the interpretability of the model in order to understand which features make contributions. Different from deep learning algorithm which is termed as a black box, interpretable machine learning is more appropriate to tackle this problem. The interpretability of a machine learning method can be divided into inherently explanation and post-hoc explanation. Inherently explanation means that the model itself is explanable. The post-hoc explanation means selecting and training the black box model, such as ensemble method or neural network, and applying interpretability methods after training\cite{55-2018Model}. The work\cite{15-wang2021A}proposes a sparsity algorithm based on post-hoc explanation in terms of corporate credit rating. The author tried to discover how to make the least amount of changes to achieve the goal of improving credit rating score, which is described as an optimization problem. Through the sparsity algorithm, enterprises can use less cost to improve their credit ratings. In addition, the study also find that the higher the credit rating, the more difficult it is for enterprises to improve their credit rating. An example of counterfactual explanation is shown in figure \ref{Example of counterfactual explanation}. By explaining the model, we can explore how to improve the credit rating score at the least cost. This idea opens up a new idea of enterprise credit rating model.

Generally speaking, the model based on neural network has relatively low requirements for Feature Engineering. It takes into account the characteristics of time series changes of rating information and the relationship between enterprises. The qualitative information that is difficult to capture by machine learning model is integrated. It also alleviates the problem of uneven data distribution. This kind of model has replaced the previous statistical model and machine learning model year by year and become the mainstream of credit rating. However, neural network model depends more on a large number of datasets, and its interpretability is not high. There is still considerable development potential in the future.

\section{Conclusion and future work}



As an inevitable product of economic development, corporate credit rating model has attracted more and more attention in recent years. By summarizing the previous literature, this paper systematically analyzes the development process of corporate credit rating. This paper deeply introduces the credit rating method from three aspects: the traditional statistical model, the CCR model based on machine learning and the CCR model based on neural network.

At present, the credit rating model has been very rich and the field of enterprise credit rating has made rapid development. The cost of time and capital has been greatly reduced. The proposal of new methods has broken the monopoly threshold of rating agencies, and small and micro enterprises can also obtain the rating score. In addition, with the introduction of neural network method, the rating accuracy has been significantly improved, and the subjectivity of analysts in the model has also been reduced.

However, there are still many problems to be solved in the field of credit rating. Micro enterprises have difficulty in rating due to the lack of corresponding data. In addition, most enterprises have good credit and few enterprises with low grades. The serious imbalance of data set categories has brought great suffering to the rating process. The data sets used in the existing models are built by researchers themselves, which leads to the lack of open source, unified and widely used datasets. A unified dataset is necessary to compare the performance of different models. Before the extensive application of deep learning, the traditional rating widely used a combination of qualitative and quantitative methods. However, most of the rapidly developing deep learning methods in recent years only use quantitative data for analysis. In addition, credit rating problems are often regarded as classification problems by deep learning models, ignoring the sensitivity of rating to ranking. The interpretability of enterprise credit rating model based on neural network remains to be discussed, which is a problem for financial practitioners. Figure \ref{Distribution chart of CCR} shows the distribution of enterprise ratings in different industries. Due to the large difference between different industries, the enterprise credit rating model needs to take into account the business characteristics of different enterprises. In the future, the research on enterprise credit rating will change to the research on credit enhancement. The rating results can only partially reflect the business status of enterprises. How to help enterprises improve their business ability and credit rating score is a new research perspective in this field in the future.



\section*{Acknowledgment}

Acknowledgmentssss

\bibliographystyle{ieeetr}
\bibliography{chengxi2}

\end{document}